%% file: main.tex
\pdfoutput=1

\documentclass[11pt]{article}

\usepackage[final]{acl}

\usepackage{times}
\usepackage{latexsym}

\usepackage[T1]{fontenc}

\usepackage[utf8]{inputenc}

\usepackage{microtype}

\usepackage{inconsolata}

\usepackage{graphicx}

\usepackage[utf8]{inputenc} 
\usepackage[T1]{fontenc}    
\usepackage{hyperref}       
\usepackage{url}            
\usepackage{booktabs}       
\usepackage{amsfonts}       
\usepackage{nicefrac}       
\usepackage{microtype}      
\usepackage{xcolor}         
\usepackage{colortbl}
\usepackage{multirow}
\usepackage{amsmath}
\usepackage{fontawesome}
\usepackage{svg}
\usepackage{siunitx}

\newcommand{\greenup}{\textcolor{green}{\faArrowUp}}
\newcommand{\reddown}{\textcolor{red}{\faArrowDown}}
\input{math_commands}

\input{macro}

\title{FactCG: Enhancing Fact Checkers with Graph-Based \\ Multi-Hop Data}

\author{Deren Lei\thanks{\enspace Equal contributions.} \qquad Yaxi Li\footnotemark[1] \qquad Siyao Li\footnotemark[1] \qquad Mengya Hu \qquad Rui Xu \\ \textbf{Ken Archer} \qquad \textbf{Mingyu Wang \qquad Emily Ching \qquad Alex Deng}\\
\\
Microsoft Responsible AI\\
\{\textit{\textbf{derenlei}, \textbf{yaxi.li}, \textbf{lisiyao}, yuetc, alex.deng}\}@\textit{microsoft.com}\\
}

\newcommand{\ours}{FactCG}

\begin{document}
\maketitle
\begin{abstract}

Prior research on training grounded factuality classification models to detect hallucinations in large language models (LLMs) has relied on public natural language inference (NLI) data and synthetic data. However, conventional NLI datasets are not well-suited for document-level reasoning, which is critical for detecting LLM hallucinations. Recent approaches to document-level synthetic data generation involve iteratively removing sentences from documents and annotating factuality using LLM-based prompts. While effective, this method is computationally expensive for long documents and limited by the LLM's capabilities. In this work, we analyze the differences between existing synthetic training data used in state-of-the-art models and real LLM output claims. Based on our findings, we propose a novel approach for synthetic data generation, CG2C, that leverages multi-hop reasoning on context graphs extracted from documents. Our fact checker model, \ours, demonstrates improved performance with more connected reasoning, using the same backbone models. Experiments show it even outperforms GPT-4-o on the LLM-A{\small GGRE}F{\small ACT} benchmark with much smaller model size. \footnote{\small \url{https://github.com/derenlei/FactCG}}
\end{abstract}

\section{Introduction}

\begin{figure}[t]
    \centering
    
    \includegraphics[width=0.98\linewidth, trim=0.6cm 0.8cm 0.8cm 1cm, clip]{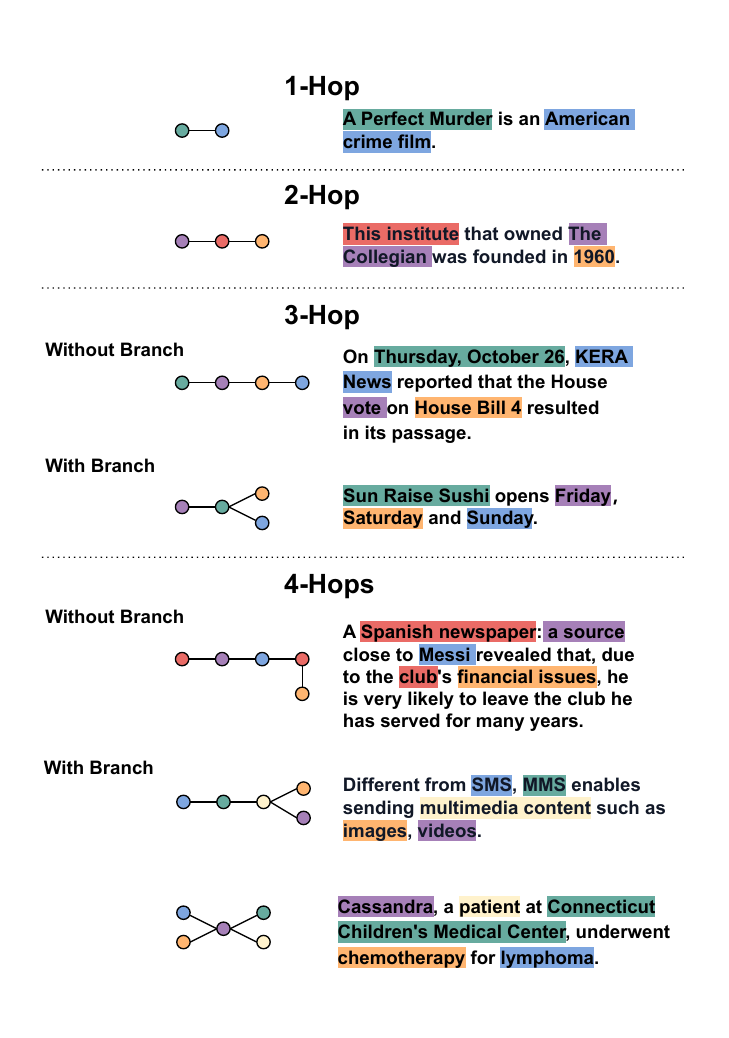}
    \vspace{-.25cm}
    \caption{\textbf{Context Graphs in LLM Responses}. Red nodes represent contextualized entities that requires further bridging for coreference resolution.}
    \vspace{-.35cm}
    \label{fig:multi-hop_example}
\end{figure}

Large language models (LLMs) have shown remarkable progress in generating fluent and coherent text across domains. They are capable of generating text conditioned on documents, leading to various applications, such as summarization \cite{zhang2024benchmarking}, retrieval augmented generation \cite{lewis2020retrieval}, code completion \cite{roziere2023code}, etc. However, one of the major challenges faced by these models is the tendency to generate hallucinated statements that are factually incorrect or not grounded in the provided document \cite{maynezetal2020faithfulness,huang2023survey,zhang2023siren}. Detecting these hallucinations is critical for providing trustworthy and responsible LLM services.

To identify hallucinations in LLM outputs (claims), recent efforts focuses on LLM-based approaches and training grounded factuality classification models. LLM-based approaches \cite{manakul2023selfcheckgpt,liu-etal-2023-g,Lei2023ChainON,dhuliawala2023chain} leveraging prompts are easy to use and benefit from the strong generalization capabilities of LLMs. However, they are not cost-effective for real application use as they often require asking LLM to generate claims multiple times \cite{manakul2023selfcheckgpt}, sending each entity of interest iteratively to identify faithfulness \cite{Lei2023ChainON}, or leveraging chain-of-thought to generate long sequence of reasoning chain per claim judgement \cite{liu-etal-2023-g, dhuliawala2023chain}. Therefore, they are mainly used as annotation tools to guide the training of smaller grounded factuality classifiers or treated as strong baselines for benchmark comparison purposes. To make online hallucinations detectors, various studies focuses on training grounded factuality classification models. Their training process usually leverages natural language inference (NLI) datasets \cite{laban-etal-2022-summac}, adapting other reasoning tasks to the NLI framework \cite{zha-etal-2023-alignscore} or synthetic data generation techniques \cite{tang-etal-2024-minicheck}. Our approach lies under this category with a focus on adapting and generating more complex data.

In this work, we investigate the patterns of existing training data and compare it to real LLM-generated claims. We notice there is a non-negligible difference in the complexity of claims and the reasoning process required for claim verification on grounding documents. We propose context graph to claim (CG2C), a synthetic data generation approach that is able to generate complex multi-hop claims in a more fine-grained and cost-effective way than past studies without any human annotation effort. It also does not leverage LLM-based approaches to generate training data labels, hence it is not bounded by LLM-based approaches' performances. We verify its effectiveness on top of state-of-the-art models by teaching them to do complex multi-hop reasoning. Experiments show that our model, FactCG, achieves state-of-the-art performance on the LLM-A{\small GGRE}F{\small ACT} \cite{tang-etal-2024-minicheck} grounded factuality benchmark compared to trained models under a billion parameters. Its performance is not bounded by GPT that is used to construct CG2C. Moreover, FactCG does more connected reasoning than leveraging dataset artifacts. 

Our contributions are three-fold:
\begin{itemize}
    \item We analyze the existing synthetic data generation for training grounded factuality models and identify the gap in claim complexity.
    \item We study the effectiveness of CG2C, a data generation approach requires no human supervision. It is able to control the complexity of generated claims with high granularity.
    \item We propose a fact checker: FactCG that achieves state-of-the-art average performance on the LLM-A{\small GGRE}F{\small ACT} benchmark compared to other models of comparable size and outperforms GPT-4-o.
\end{itemize}

\section{Related Work}
\start{Fact-Checking Methods}
There are two main approaches to detect hallucinations without altering LLM architectures. One relies on LLMs evaluating their own outputs, like SelfCheckGPT \cite{manakul2023selfcheckgpt}, or leveraging Chain-of-Thought strategies, as seen in G-Eval \cite{liu-etal-2023-g}, CoVe \cite{dhuliawala2023chain}, and CoNLI \cite{Lei2023ChainON}. The other focuses on training cost-efficient factuality classifiers. SummaC \cite{laban-etal-2022-summac} adapts NLI models for document-level factual checks, while QAFactEval \cite{fabbri-etal-2022-qafacteval} uses Question Answering (QA) metrics. FactCC \cite{kryscinski-etal-2020-evaluating} predicts summary consistency, and AlignScore \cite{zha-etal-2023-alignscore} trains on large-scale unified NLI data. MiniCheck \cite{tang-etal-2024-minicheck} generates training data using LLMs to label and outperforms previous work on human-annotated real LLM hallucination benchmark LLM-A{\small GGRE}F{\small ACT} \cite{tang-etal-2024-minicheck}. Our work continue the study on improving fact-checking models by enhancing LLM-generated data complexity and controllability.

\start{Fact-Checking Data} Several studies focus on creating datasets for hallucination detection, either for training or benchmarking. A{\small GGRE}F{\small ACT} \cite{tang-etal-2023-understanding} evaluates factual consistency in summarization (CNN/DM, XSum), while TofuEval \cite{tang-etal-2024-tofueval} targets dialogue summarization. WiCE \cite{kamoi-etal-2023-wice} is a textual entailment dataset based on Wikipedia claims. REVEAL \cite{jacovi-etal-2024-chain} benchmarks Chain-of-Thought reasoning in open-domain QA, and ClaimVerify \cite{liu-etal-2023-evaluating} evaluates correctness and groundedness in generative search engines' responses. FactCheck \cite{Wang2023FactcheckGPTEF} constructs a document-level factuality benchmark, while EXPERTQA \cite{malaviya-etal-2024-expertqa} focuses on long-form QA with verified attributions. LFQA \cite{chen-etal-2024-fintextqa} assesses LLM-generated sentences for factual support by the documents that are either retrieved by humans, models, or randomly selected. RAGTruth \cite{niu-etal-2024-ragtruth} analyzes word-level hallucinations in various domains and tasks within the standard retrieval-augmented generation frameworks for LLM applications. HaluEval \cite{HaluEval} is an LLM evaluation benchmark with synthetically generated hallucination data.

\start{Multi-hop Reasoning}
Multi-hop reasoning refers to inference over multiple steps. Various studies focused on generating QA datasets where multi-hop reasoning on documents is required to get the answer, such as HotpotQA \cite{yang-etal-2018-hotpotqa}, 2WikiMultihopQA \cite{ho-etal-2020-constructing}, Musique \cite{trivedi2022musique}. However, various studies show that models often exploit dataset artifacts to produce correct answers without connecting information across multiple supporting facts even when multi-fact reasoning is required \cite{jiang2019avoiding,chen2019understanding,min2019compositional}, namely disconnected reasoning (DiRe) \cite{trivedi-etal-2020-multihop}.


\start{LLM-based Graph Construction}
Recent literatures have shown the effectiveness of LLM in graph construction tasks. \citeauthor{trajanoska2023enhancing} leverages LLM to enhance knowledge graph construction from unstructured text. KG-LLM \cite{yao2023exploring} explored the LLM-base approach to complete the knowledge graph. Graph RAG \cite{edge2024local} utilized LLM generated graphs to improve the RAG system on global questions that target the entire text corpus.

\section{Problem Formulation}
Given a document $doc$ and a claim $c$, we consider $c$ to be grounded in $doc$ if a generic reader would affirm the statement "According to $doc$, $c$ is true".
Conversely, $c$ is ungrounded with respect to $doc$ if it conflicts with or cannot be verified against $doc$. Our objective is to model a function to detect ungrounded hallucinations in LLM-generated claims against documents at the sentence-level.

We define grounded as 1 and ungrounded as 0. $\gM$ is a classifier that first maps each $\langle doc, c \rangle$ pair into a confidence score of judging $c$ is grounded in $doc$, formally defined as $\gM(doc, c) \in [0,1]$. 
We then use a threshold to convert confidence score into a binary prediction
\begin{equation}
y  = P(\gM(doc, c) \geq \theta \mid doc)
\end{equation}
where $\theta$ is the threshold we can select.



As most recent grounded factuality detection industrial applications now support dynamic threshold tuning \cite{google-grounding, amazon-grounding}, we consider the threshold to be dynamically adjustable to its optimal level for per-scenario detection and follow the past work \cite{laban-etal-2022-summac, zha-etal-2023-alignscore, tang-etal-2023-understanding, tang-etal-2024-tofueval}.

\section{Analysis on Multi-Hops in Claims}
\label{sec:poc}
In this section, we conduct a proof of concept experiment on the multi-hop in LLM-generated claims. We try to answer the following two research questions (RQ):

\smallskip
\noindent
\textbf{RQ1}: To what extent do LLM-generated claims, based on grounding documents, require multi-hop reasoning for fact verification?

\smallskip
\noindent
\textbf{RQ2}: To what extent do current state-of-the-art synthetic claims, based on grounding documents, require multi-hop reasoning for fact verification?

\subsection{Analysis Setup} \label{Section: Analysis Setup}
\start{LLM Claims} To answer RQ1, we select the RAGTruth \cite{niu-etal-2024-ragtruth} validation dataset from the LLM-A{\small GGRE}F{\small ACT} benchmark \cite{tang-etal-2024-minicheck} for analysis, as it contains three recognized tasks with RAG settings: Question Answering (QA), Data-to-text Writing (Data2Text), and News Summarization (Summ). It also leverages six different LLMs to generate completion responses. We analyze each generation task separately. To preserve the original LLM completion format, we do not further decompose the claim into atomic claims. For each task, we deduplicate the documents in the entire set and randomly select 500 documents and their corresponding document-claim pairs for analysis to reduce the computational cost. 


\start{Synthetic Claims}  To answer RQ2, we select MiniCheck's D2C synthetic claims \cite{tang-etal-2024-minicheck} for analysis as the classifier model trained on it achieves the best result on LLM-A{\small GGRE}F{\small ACT}. Same as RAGTruth claims, we do not further process the D2C claims and analyze them on a sentence-level basis as provided. We conduct analysis on the entire dataset without random sampling.

\start{Multi-hop Context Sub-graph Extraction} As the LLMs generate claims conditioned on grounding documents, they are not always decontextualized and may contain implicit reasoning chains. To extract the complete reasoning chain of each claim for both datasets, we leverage LLMs in the following steps:

Firstly, we construct the context graph $\mathcal{G}$ from the grounding document $doc$. 
We follow GraphRag's \cite{edge2024local} prompt-based approach to extract $\langle entity, entity, relation \rangle$ triples from $doc$, where the $relation$ is a short sentence that describes how the two entities are connected based on the context of the document, which is non-directional. We extract all triples and further group them so that each triple shares at least one entity within its cluster.

Secondly, we find the context sub-graph $\mathcal{G}_c$ for each claim $c$ by selecting the triples related to it. The $\mathcal{G}_c$ contains the information in $c$ that requires verification against $doc$.


Thirdly, we count the number of connected triples in each $\mathcal{G}_c$ as the hop number for $c$. If $\mathcal{G}_c$ has multiple sub-graphs, we count the hops of the largest sub-graph as the hop number.

We calculate the distribution of hops in the reasoning chains extracted for each dataset.

\subsection{Experiment Settings}
 We use GPT-4-o for context graph extraction and context sub-graph mapping with claims. We remove data with ill-formatted GPT outputs in both steps to simplify the post-processing. This experiment is only conducted on positive samples in every dataset because ungrounded claims may take any form and their reasoning chain may not exist in the source document's graph.  

\subsection{Results and Analysis}
We show multi-hop extraction results in Table \ref{tb:poc}. Synthetic data in MiniCheck's D2C dataset lean towards claims with fewer hops, while LLM-generated claims in RAGTruth tend to have higher ratios of 2-hop and 3-hop claims. We even found non-negligible amounts of claims with 4 or more reasoning hops in RAGTruth across all tasks, which are very limited in D2C. We believe bridging this gap is a non-trivial task and we hypothesize that teaching the model to learn multi-hop inference may improve the model's overall capability for hallucination detection.


\begin{table}[htbp]
\centering
\resizebox{\columnwidth}{!}{
\begin{tabular}{c|ccc|c}
\hline
\textbf{Hops} & \textbf{Data2Text} & \textbf{Summ} & \textbf{QA} & \textbf{D2C} \\ \hline
1-hop & 42.6\% & 42.7\% & 51.1\% & 71.0\% \\ \hline
2-hop & 32.1\% & 36.6\% & 32.9\% & 22.6\% \\ \hline
3-hop & 17.3\% & 15.2\% & 10.9\% & 5.3\%  \\ \hline
4-hop & 6.5\%  & 3.4\%  & 3.3\%  & 1.1\%  \\ \hline
$\geq$ 5-hop & 1.4\%  & 2.0\%  & 1.9\%  & 0.0\%  \\ \hline
\end{tabular}
}
\caption{\textbf{Claim Hop Analysis.} Hop distribution of real LLM-generated claims in RAGTruth data2text, summarization and question-answering split versus MiniCheck D2C synthetic claims.}
\label{tb:poc}
\end{table}

\section{Improving Model with Context Graph to Multi-hop Claim (CG2C)}
\begin{figure*}[t]
    \centering
    \includegraphics[width=0.98\linewidth, trim=0cm 0cm 0cm 0cm, clip]
    {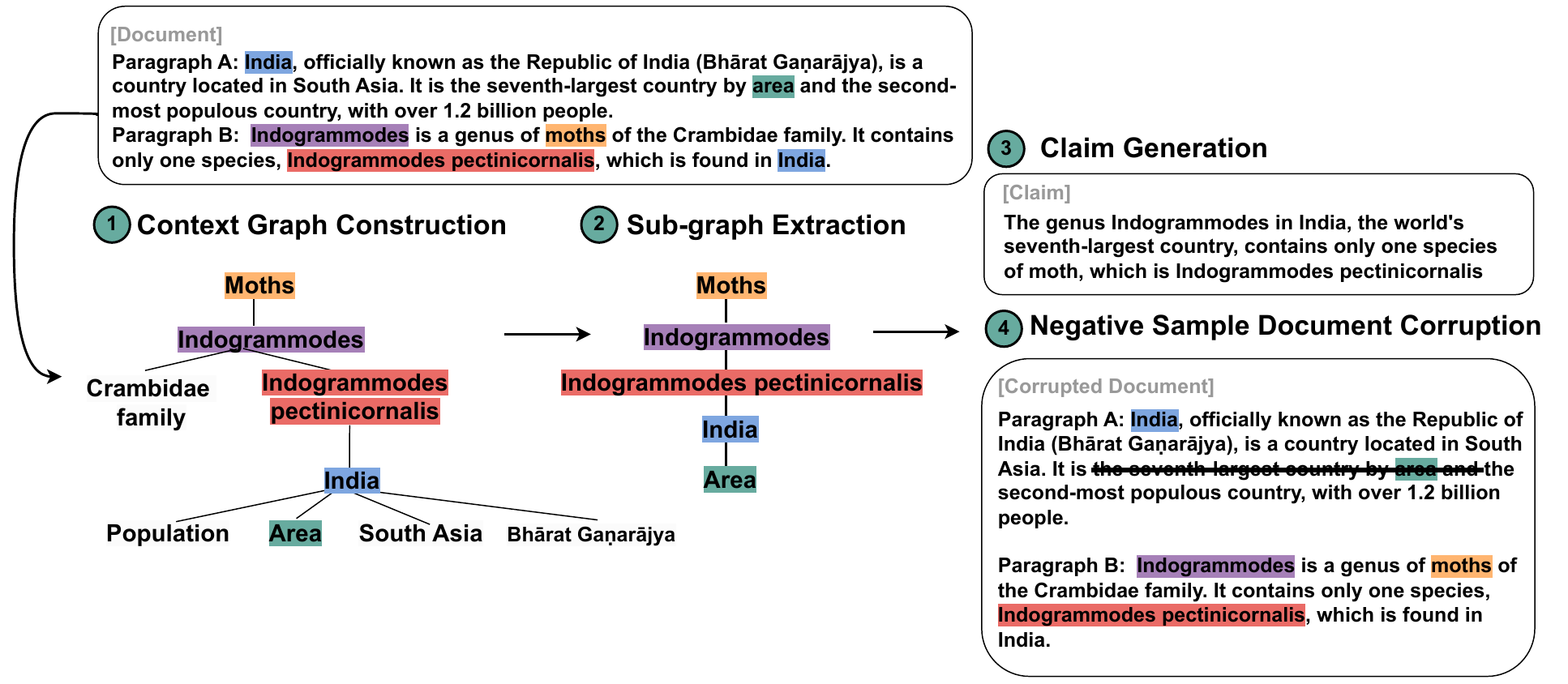}
    \caption{\textbf{CG2C from Document.} To generate synthetic data with documents only, first we construct context graph $\mathcal{G}$ from document $doc$. Second we extract sub-graph $\mathcal{G}_c$ from the context graph as multi-hop. Third we generate claim $c$ with context $\mathcal{G}_c$. To corrupt the document we remove a random relation between entities within $\mathcal{G}_c$ to get $doc_{neg}$. Finally, we get positive sample $\langle doc, c \rangle$ and negative sample $\langle doc_{neg}, c \rangle$. For CG2C-MHQA, we get $c$ from $\langle q, ans \rangle$ and use $\langle \mathcal{G}, c \rangle$ to get $\mathcal{G}_c$ instead of step 2 and 3 mentioned above.}
    \label{fig_framework}
\end{figure*}

In this section, we study whether we can improve the state-of-the-art grounded factuality models with synthetically generated data via context graph to multi-hop claims (CG2C). Specifically, the study follows two steps, each addressing a research question:

\smallskip
\noindent
\textbf{RQ3}: Can we improve grounded factuality models by generating synthetic data leveraging large-scale public multi-hop datasets?

\noindent
\textbf{RQ4}: Can we improve grounded factuality models by generating synthetic multi-hop data leveraging documents only?

\subsection{CG2C from Multi-hop QA}
\label{sec: cg2c from qa}
In this section, we study RQ3. We leverage publicly available multi-hop question answering (MHQA) datasets to synthetically construct grounded factuality data for model training, namely CG2C-MHQA. Specifically, we use HotpotQA \cite{yang-etal-2018-hotpotqa} and Musique \cite{trivedi2022musique}, where each data sample contains a source document $doc$, a question $q$, an answer $ans$ and supporting sentences $doc_{support}$, which is a subset of $doc$ that provides evidence for the answer. We leverage LLMs to convert $\langle ans, q \rangle$ to a single declarative statement. We define the converted statement as claim $c$.

\start{HotpotQA} We leverage the train-hard split in HotpotQA to ensure each data sample requires complex multi-hop reasoning across multiple paragraphs. We treat the $\langle doc, c \rangle$ pair as a positive sample. We then randomly select positive samples to corrupt them into negative samples by following the steps below. Firstly, since $doc_{support}$ is provided, we extract $\mathcal{G}_s$ from $doc_{support}$ and then leverage the $\langle \mathcal{G}_s, c \rangle$ pair to get a context sub-graph $\mathcal{G}_c$, using the same strategy described in Section \ref{Section: Analysis Setup}. Secondly, from the $\mathcal{G}_c$, we randomly select a triple and leverage LLMs to remove the relation between entities in the triple from $doc$ to form $doc_{neg}$. Finally, the $\langle doc_{neg}, c \rangle$ pair forms a negative sample. Inspired by  \citeauthor{nie-etal-2020-adversarial} (\citeyear{nie-etal-2020-adversarial}), we further reduce the dataset size by leveraging RoBERTa-large trained on the MultiNLI dataset, \cite{williams-etal-2018-broad} to select difficult samples. As the NLI model predicts each sample from $\{\text{Entailment}, \text{Contradiction}, \text{Neutral}\}$, we define a mapping function $f(\cdot)$ for the model output $y$ as:

\begin{small}
\[
f(y) = \begin{cases} 
\text{Grounded} & \text{if } y = \text{Entailment}, \\
\text{Ungrounded} & \text{if } y \in \{\text{Contradiction}, \text{Neutral}\}
\end{cases}
\]
\end{small}

We remove data that $\text{RoBERTa}_\text{MNLI}$ \footnote{https://huggingface.co/FacebookAI/roberta-large-mnli/tree/main} predicts correctly. 





\start{Musique} We leverage 3-hop and 4-hop data from the Musique dataset to diversify from HotpotQA where most of the data is 2-hop. The $\langle doc, c \rangle$ pairs with $c$ converted from unanswerable $\langle q, ans \rangle$ pairs are treated as negative samples and answerable pairs are treated as positive samples. Similar to HotpotQA, we remove data that $\text{RoBERTa}_\text{MNLI}$ predicts incorrectly. 


\start{Merge} We merge synthetically constructed data from HotpotQA and Musique into CG2C-MHQA.

\subsection{CG2C from Documents}
\label{sec: cg2c from doc}
In this section, we study RQ4. We further investigate the possibility of leveraging any document to directly generate synthetic training data, namely, CG2C-Doc. 
We propose the following steps to generate harder and more diverse data with fine-grained control:

Firstly, we extract a context graph from the given document $doc$ following the same approach as in Section \ref{Section: Analysis Setup}.
Secondly, after filtering out cyclic graphs, we programmatically extract sub-graphs with a specific number of hops and shapes. The types of sub-graphs we extract are shown in Figure \ref{fig:multi-hop_example}.
Thirdly, for each generated sub-graph, we leverage LLMs to generate a claim $c$ which includes all graph nodes. The $\langle doc, c \rangle$ pair forms a positive data sample.
Lastly, we randomly extract a relation from the sub-graph, and request LLMs to remove it from the $doc$ to create the document for negative sample $\langle doc_{neg}, c \rangle$ . 

Unlike previous work \cite{tang-etal-2024-minicheck}, where LLMs are used directly to decide the synthetic samples' labels, our proposed data generation strategy avoids involving LLMs in label generation, hence is less bounded by the LLM prompt-based approach's performance. We can also easily generate more complex data (3-hop or 4-hop samples) for each document by controlling the sub-graph we extract. 

For a fair comparison with the current state-of-the-art MiniCheck \cite{tang-etal-2024-minicheck}, we use the same documents from its D2C dataset to generate CG2C-Doc. We only use data from 3-hop and 4-hop sub-graphs without branches for training because we find them empirically better. We analyze this result in the Ablation Study Section \ref{sec:ablation}.

\subsection{Proposed Fack Checker: FactCG}
\start{Model Backbones and Synthetic Data Baseline}
We follow MiniCheck \cite{tang-etal-2024-minicheck} training data and the two-stage training process as the synthetic data training baseline.
We also select the same model backbones as MiniCheck for easier experiment comparison, namely RoBERTa-large (RBT) \cite{liu2019roberta}, DeBERTa-v3-large (DBT) \cite{he2023debertav} and Flan-T5-large (FT5) \cite{chung2024scaling}. We add our synthetically generated data into the MiniCheck training pipeline and perform an apple-to-apple comparison.

\start{Model Input and Output} We do conventional sequence pair classification for FactCG-RBT. For FactCG-FT5, we use the FLAN instruction \cite{chung2024scaling} for the ANLI \cite{nie-etal-2020-adversarial} dataset as the input template. At inference time, we leverage "Yes" and "No" logits from the language model head as the final classification logits. For FactCG-DBT, we use the same FLAN input template as FactCG-FT5 and perform sentence classification training. We find it to be empirically effective. We apply the softmax function to transform the model's raw logits then use threshold $\theta$ to get final binary output $y$.

\start{Chunking at Inference Time} Following recent works \cite{tang-etal-2024-minicheck, zha-etal-2023-alignscore}, we chunk the document into sequential chunks at sentence boundaries during inference time. We set the target chunk size for \ours-RBT, \ours-DBT, and \ours-FT5 to 400, 550, and 550 tokens respectively. The highest score across chunks will be treated as the final score.


\section{Experiments}
In this section, we compare FactCG models with other approaches on grounded factuality datasets.

\subsection{Datasets}
We leverage LLM-A{\small GGRE}F{\small ACT} benchmark \cite{tang-etal-2024-minicheck}, which consists of real hallucinations generated from recent language models. It includes 11 datasets, namely A{\small GGRE}F{\small ACT}-CNN, A{\small GGRE}F{\small ACT}-XSum \cite{tang-etal-2023-understanding}, TofuEval-MediaS, TofuEval-MeetB\cite{tang-etal-2024-tofueval}, WiCE \cite{kamoi-etal-2023-wice}, REVEAL \cite{jacovi-etal-2024-chain}, ClaimVerify \cite{liu-etal-2023-evaluating}, FactCheck \cite{Wang2023FactcheckGPTEF}, EXPERTQA \cite{malaviya-etal-2024-expertqa}, LFQA \cite{chen-etal-2024-fintextqa} and RAGTruth \cite{niu-etal-2024-ragtruth}. 

\subsection{Experiment Setup}

\start{Synthetic Data Construction} We use GPT-4-o for the CG2C data generation pipeline described in Section \ref{sec: cg2c from qa} and Section \ref{sec: cg2c from doc}. The number of synthetic data we generated from each source is shown in Table \ref{tb:data count}.  

\begin{table}[htbp]
\centering
\resizebox{\columnwidth}{!}{
\begin{tabular}{cccc}
\hline
\multirow{2}{*}{\textbf{Dataset}} & \multicolumn{2}{c}{\textbf{CG2C-MHQA}} & \multirow{2}{*}{\textbf{CG2C-Doc}}\\
 & \textbf{HotpotQA} & \textbf{Musique} & \\ \hline
Positive & 6734 & 5385 & 1034 \\ \hline
Negative & 1479 & 1048 & 1034 \\ \hline
Total & 8213 & 6433 & 2068 \\ \hline
\end{tabular}

}
\caption{\textbf{CG2C Dataset.} Number of synthetic data generated from different sources.}
\label{tb:data count}
\end{table}

\begin{table*}[ht]
\centering
\caption{\textbf{LLM-A{\small GGRE}F{\small ACT} Results with Threshold Tuning.} Evaluation using BAcc (\%). GPT results are added for reference. Highest score (except GPT) for each dataset is highlighted with dark green and second highest score highlighted with light green.}
\resizebox{2\columnwidth}{!}{
\begin{tabular}{lcccccccccccc}
\toprule
\multirow{2}{*}{\textbf{Model}} & \multicolumn{2}{c}{\textbf{A{\small GGRE}F{\small ACT}}} & \multicolumn{2}{c}{\textbf{TofuEval}} & \multirow{2}{*}{\textbf{WiCE}} & \multirow{2}{*}{\textbf{REVEAL}} & \textbf{Claim} & \textbf{Fact} & \textbf{Expert} & \multirow{2}{*}{\textbf{LFQA}} & \textbf{RAG} & \multirow{2}{*}{\textbf{AVG}} \\
 \textbf{} & \textbf{CNN} & \textbf{XSum} & \textbf{MediaS} & \textbf{MeetB} & \textbf{} & \textbf{} & \textbf{Verify} & \textbf{Check} & \textbf{QA} & \textbf{} & \textbf{Truth} & \textbf{} \\
\midrule
GPT-4o-2024-05-13 & 68.1 &  76.8 & 71.4 & 79.8 & 78.5 & 86.5 & 69.0 & 77.5 & 59.6 & 83.6 &  84.3 & 75.9 \\
\midrule
\midrule
SummaC-ZS & 62.7 & 67.6 & 68.5 & 71.1 & 63.9 & 86.3 & 70.0 & 75.2 & 58.1 & 81.9 & 65.1 & 70.0 \\
SummaC-Conv & 67.7 & 64.6 & 68.4 & 69.2 & 65.2 & 85.2 & 71.3 & 74.2 & 57.8 & 79.0 & 65.1 & 69.8 \\
AlignScore & 63.5 & 69.4 & 71.8 & 71.8 & 66.2 & 85.3 & 72.9 & 76.6 & \cellcolor{lightgreen} 59.2 & 86.0 & 74.3 & 72.5 \\
AlignScore-DBT & 65.4 & 72.6 & 70.2 & 76.1 & 69.9 & 86.4 & 74.1 & 76.9 & 57.8 & 83.7 & 74.3 & 73.4 \\
MiniCheck-RBT & 65.4 & 70.1 & 71.1 & 74.9 & 73.7 &  88.0 & 77.1 & \cellcolor{lightgreen} 77.5 & 59.0 & 84.1 & 77.3 & 74.4 \\
MiniCheck-DBT & 63.7 & \cellcolor{lightgreen} 74.6 & 68.9 & 70.6 & 76.5 & 87.4 & 75.6 & 75.9 & 58.8 & 84.2 & 78.9 & 74.1 \\
MiniCheck-FT5 & 70.1 & \cellcolor{darkgreen} 74.7 & \cellcolor{lightgreen} 73.1 & \cellcolor{lightgreen} 76.2 & 76.0 & 86.6 & 74.7 & 76.8 & 58.6 & 85.7 & 78.5 & 75.5 \\ 
\midrule
\ours-RBT & \cellcolor{lightgreen} 70.6 & 72.2 & 71.2 & 75.3 & 72.1 & \cellcolor{lightgreen} 89.9 &  78.2 &  76.7 & 58.7 &  \cellcolor{lightgreen} 86.6 & 78.1 & 75.4 \\
\ours-DBT & \cellcolor{darkgreen} 73.8 & \cellcolor{lightgreen} 74.6 & \cellcolor{darkgreen} 73.7 & 73.9 & \cellcolor{darkgreen} 80.9 & \cellcolor{darkgreen} 90.0 & \cellcolor{lightgreen} 78.3 & 75.1 & \cellcolor{darkgreen} 59.4 & \cellcolor{lightgreen}86.6 & \cellcolor{darkgreen}82.6 & \cellcolor{darkgreen}77.2 \\
\ours-FT5 & 69.8 & 73.8 & \cellcolor{darkgreen} 73.7 & \cellcolor{darkgreen} 76.5 & \cellcolor{lightgreen} 77.8 & 89.6 & \cellcolor{darkgreen} 79.3 & \cellcolor{darkgreen} 77.6 &  59.1 & \cellcolor{darkgreen} 87.7 & \cellcolor{lightgreen} 79.3 & \cellcolor{lightgreen} 76.7 \\
\bottomrule
\end{tabular}
}
\label{tab:benchmark}
\end{table*}

\begin{table*}[ht]
\centering
\caption{\textbf{Ablation Study on CG2C.} Evaluation using BAcc (\%). Removing both CG2C-Doc and CG2C-MHQA results in the lowest performance. Removing either one while retaining the other leads to a smaller performance drop compared to removing both datasets. Consistent observations on three backbone models.}
\resizebox{2\columnwidth}{!}{
\begin{tabular}{lcccccccccccc}
\toprule[1.0pt]
\multirow{2}{*}{\textbf{Model}} & \multicolumn{2}{c}{\textbf{A{\small GGRE}F{\small ACT}}} & \multicolumn{2}{c}{\textbf{TofuEval}} & \multirow{2}{*}{\textbf{WiCE}} & \multirow{2}{*}{\textbf{REVEAL}} & \textbf{Claim} & \textbf{Fact} & \textbf{Expert} & \multirow{2}{*}{\textbf{LFQA}} & \textbf{RAG} & \multirow{2}{*}{\textbf{AVG}} \\
 \textbf{} & \textbf{CNN} & \textbf{XSum} & \textbf{MediaS} & \textbf{MeetB} & \textbf{} & \textbf{} & \textbf{Verify} & \textbf{Check} & \textbf{QA} & \textbf{} & \textbf{Truth} & \textbf{} \\
\toprule

\ours-RBT \\
$-$ CG2C-Doc & -1.94 \reddown & +0.05 \greenup  & -1.24 \reddown & -1.10 \reddown & -3.08 \reddown & -0.01 
\reddown & -0.39 \reddown & -0.23 \reddown & +0.62 \greenup & -1.00 \reddown & +0.27 \greenup & -0.73 \reddown \\
$-$ CG2C-MHQA & -4.72 \reddown & +0.90 \greenup & -0.34 \reddown & -0.22 \reddown & +0.46 \greenup & -1.45 \reddown & -1.41 \reddown & +0.44 \greenup & -0.22 \reddown & +0.04 \greenup & -1.17 \reddown & -0.70 \reddown \\
$-$ CG2C-Both & -8.06 \reddown & -0.96 \reddown & -1.85 \reddown & +0.10 \greenup & -4.37 \reddown & -2.29 \reddown & -2.36 \reddown & +1.35 \greenup & +0.09 \greenup & 
-0.78 \reddown & -1.23 \reddown & -1.85 \reddown \\
\midrule
\ours-DBT \\
$-$ CG2C-Doc & -1.40\reddown & +1.14\greenup & +0.72\greenup & +2.13\greenup & -2.74\reddown & -3.08\reddown & -1.40\reddown & +0.12\greenup & -0.9\reddown & -0.16\reddown & -0.35\reddown & -0.54\reddown \\
$-$ CG2C-MHQA & -1.95\reddown & +1.60\greenup & -1.33\reddown & +2.04\greenup & -4.58\reddown & -0.30\reddown & +0.46\greenup & +2.70\greenup & -0.35\reddown & +1.74\greenup & -1.02\reddown & -0.09\reddown \\

$-$ CG2C-Both &  -10.12\reddown & -0.05\reddown & -4.79\reddown & -3.32\reddown & -4.36\reddown & -2.67\reddown & -2.68\reddown & +0.85\greenup & -0.60\reddown & -2.40\reddown & -3.71\reddown & -3.08\reddown \\

\midrule

\ours-FT5 \\
$-$ CG2C-Doc & -5.79 \reddown  & -1.91 \reddown & -2.39 \reddown  &  +0.86 \greenup &  +0.91 \greenup & -0.66\reddown  &  -0.47 \reddown  & -0.89 \reddown  & +0.63 \greenup  & -1.06 \reddown & -0.97 \reddown  & -1.07 \reddown  \\
$-$ CG2C-MHQA & -1.67 \reddown & -2.13 \reddown & -1.88 \reddown & -1.57 \reddown & -1.82 \reddown & -0.38 \reddown & -0.99 \reddown  & +0.39 \greenup & -0.07 \reddown & +0.50 \greenup & -0.31 \reddown & -0.90 \reddown \\
$-$ CG2C-Both & -0.15 \reddown & -3.99 \reddown & -2.69 \reddown & +0.47 \greenup & -3.44 \reddown & -2.40 \reddown & -1.72 \reddown & -0.15 \reddown & +1.06 \greenup & -0.52 \reddown & -0.34 \reddown & -1.26 \reddown \\
\bottomrule
\end{tabular}
}
\label{tb:ablation}
\end{table*}

\start{Training Details} In this section, we provide the training and hyperparameter details. We employed standard cross-entropy loss to train all the models. Unless otherwise specified, we utilized the default parameters of the backbone models. 

For \textbf{\ours-RBT}, following \citeauthor{tang-etal-2024-minicheck}(\citeyear{tang-etal-2024-minicheck}), we begin by fine-tuning the RoBERTa-large from AlignScore \cite{zha-etal-2023-alignscore}. We load both the base model and the binary classification layer. For the first stage, we used a batch size of 2, an accumulated step of 8, and a learning-rate of 1e-5 to fine-tune the model for 1 epoch on the MiniCheck's C2D data along with our CG2C-MHQA data.

For \textbf{\ours-DBT} and \textbf{\ours-FT5}, we start from DeBERTa-v3-large\footnote{https://huggingface.co/microsoft/deberta-v3-large} and Flan-T5-large\footnote{https://huggingface.co/google/flan-t5-large} respectively. For the first stage, we used a batch size of 2 and an accumulated step of 8 to fine-tune the models for 1 epoch on MiniCheck's ANLI subset and C2D along with our CG2C-MHQA. We used learning rates of 1e-5 and 5e-5 respectively.

Same as MiniCheck, we then fine-tune these models for a second stage for 1 epoch on MiniCheck's D2C, along with our CG2C-Doc datasets. Each model's training settings remain the same as in the first stage.

\start{Dynamic Threshold} We adjust the threshold for each LLM-A{\small GGRE}F{\small ACT} dataset to two decimal places, optimizing for the best performance on the validation set. We investigate how models behave when they are at their best performance on each dataset. Unlike \cite{tang-etal-2024-minicheck}, we observe that building a unified threshold system leads to different results (shown in Appendix \ref{sec: threshold0.5}), potentially underestimating the true potential of the model. We leave data calibration for a unified threshold as future work.

\start{Evaluation Metrics} Following previous works, we evaluate performance with balanced accuracy (BAcc) \cite{Brodersen2010TheBA,laban-etal-2022-summac}. $\quad \text{BAcc} = \frac{1}{2} \left( \frac{\text{TP}}{\text{TP} + \text{FN}} + \frac{\text{TN}}{\text{TN} + \text{FP}} \right),\quad$ where TP, TN, FP, and FN stand for true positives, true negatives, false positives and false negatives.

\start{Baseline Comparison}
We compare FactCG models with recent state-of-the-art fact-checking models, SummaC-ZS, SummaC-Conv \cite{laban-etal-2022-summac}, AlignScore \cite{zha-etal-2023-alignscore}, MiniCheck \cite{tang-etal-2024-minicheck} with RoBERTa-large and DeBERTa-v3-large backbones. Since AlignScore was originally trained using the RoBERTa-large backbone, we additionally trained an AlignScore variant with the DeBERTa-v3-large backbone, referred to as AlignScore-DBT, to ensure a fair apples-to-apples comparison during benchmarking. 

\subsection{Evaluation on Benchmark}
We show results on LLM-A{\small GGRE}F{\small ACT} in Table \ref{tab:benchmark}. Each experiment was ran only once due to its cost. Scores highlighted with dark green and light green are the best and second best performances for each dataset except GPT. We add GPT results for reference because both CG2C and MiniCheck use GPT to synthesize data. On average, our \ours s outperform previous solutions with the same backbones. Our \ours-DBT achieves the state-of-the-art, even outperforming GPT-4. We believe it is because our methods do not directly rely on LLMs to produce the label, so it is possible for \ours~ to surpass the boundary established by LLMs. Our \ours-DBT, although much smaller than its FT5 counterpart, reaches similar (or even higher) performance. It may be because RoBERTa is encoder-based and suitable for classification tasks in nature.

\subsection{Ablation Study}
\label{sec:ablation}

As we achieve better performance than other state-of-the-art models with similar parameter size on LLM-A{\small GGRE}F{\small ACT} with a dynamic threshold, to better answer RQ3 and RQ4, we conduct an ablation study to see the performance difference when dropping CG2c-Doc or CG2C-MHQA or both. If both are dropped, we train stage 1 for 2 epochs same as MiniCheck \cite{tang-etal-2024-minicheck} to ensure better model convergence. Results are shown in Table \ref{tb:ablation}. Dropping both CG2C-Doc and CG2C-MHQA leads to lowest performances on each backbone model. In contrast, removing either one leads to a less performance drop comparing to remove both, which means adding either CG2C data can improve the model quality. This demonstrates the effectiveness of both synthetic generation approaches and answers RQ3 and RQ4 that both CG2C-MHQA and CG2C-Doc helps improving model performance.

\begin{table}[t]
\centering
\caption{\textbf{Ablation Study on Sub-graph Shapes.} CG2C without branches achieves higher BACC on A{\small GGRE}F{\small ACT} than those with branches.}
\small
\begin{tabular}{llc}
\toprule
\textbf{Model} & \textbf{Context Sub-Graph Type} & \textbf{AVG} \\
\midrule
\multirow{2}{*}{\textbf{FactCG-RBT}} & CG2C with Branches & 74.7 \\
                                     & CG2C without Branches & \textbf{75.4} \\
\midrule
\multirow{2}{*}{\textbf{FactCG-DBT}} & CG2C with Branches & 77.1 \\
                                     & CG2C without Branches & \textbf{77.2} \\
\midrule
\multirow{2}{*}{\textbf{FactCG-FT5}} & CG2C with Branches & 76.0 \\
                                     & CG2C without Branches & \textbf{76.7} \\
\bottomrule
\end{tabular}

\label{tb:ablation-branch}
\end{table}

In addition, as shown in Table \ref{tb:ablation-branch}, we observe that leveraging multi-hop data generated from context sub-graphs with branches performs worse than generate all data without branches. This is potentially caused by they can be further decomposed into multiple sub-graph without branches. Hence converting claims to simpler sub-claims with less hops and reduced reasoning complexity.



\section{Evaluation on Connected Reasoning}
As QA models suffer from disconnected reasoning, they leverage exploited dataset artifacts to produce correct answers instead of connecting information across multi-hops. We conduct a similar connected reasoning (CoRe) analysis on the multi-hop reasoning chain of trained grounded factuality models similar to \citeauthor{trivedi-etal-2020-multihop} (\citeyear{trivedi-etal-2020-multihop}). We would like to answer the research question:

\smallskip
\noindent
\textbf{RQ5}: Does \ours~has more connected reasoning than models not trained on CG2C datasets?

\subsection{CoRe Dataset Construction}
\label{sec:core}
We leverage WiCE \cite{kamoi-etal-2023-wice} in the LLM-A{\small GGRE}F{\small ACT} benchmark as it provides the document-claim-evidence triples in the forms of $(doc_i, c_i, \mathcal{E}_i)$
where $\mathcal{E}_i = \{E_{i1}, E_{i2}, .., E_{ij}, ...\}$. Each evidence $E_{ij}$ is a unique minimum independent sentence set $E_{ij} \subset doc_i$ sufficient to make a grounded factuality judgement. $E_{ij}=\{e_{ij}^1,e_{ij}^2, ..., e_{ij}^k,...\}$. $E_{ij}$ may share the same evidence sentences. We define $E_{i}^*$ as the union of all evidence sets such that $E_{ij} \subset E_{i}^*$.

We filter positive cases in the WiCE validation and test sets that contain more than one evidence sentence in each $E_{ij}$. We treat these as the positive set that requires connected reasoning across sentences with sufficient context. 


To synthetically construct negative cases, for each positive case, we randomly select evidence sentences $e_{i}^-$ in $E_{i}^*$ and remove their appearance in the grounding document such that each $E_{ij}$ has one evidence removed. In this case, the claim cannot be verified with the corrupted document through any $E_{ij}$. We treat this negative set as minimum disconnected reasoning set with insufficient context.

For each $c_i$ we have a pair of positive and negative data $\langle doc_i, c_i \rangle$ and $\langle doc_i\backslash e_i^-,c_i \rangle$, where $i \in I$ and $|I|$ is the total size of pairs in the constructed dataset. For a grounded factuality model $\gM$, we define the accuracy via connected reasoning as:

\begin{small} 
\begin{multline}
\text{Accuracy}_\text{CoRe} = \\
 \frac{\sum_i(\gM(doc_i,c_i) \geq \theta) \cap (\gM(doc_i\backslash e_i^-,c_i) < \theta)}{|I|}
\end{multline}
\end{small}

We also define the proportion of correct predictions that were made via connected reasoning as: 
\begin{small} 
\begin{multline}
\text{Precision}_\text{CoRe} = \\
\sum_i \frac{(\gM(doc_i,c_i) \geq \theta) \cap (\gM(doc_i\backslash e_i^-,c_i) < \theta)}{(\gM(doc_i,c_i)\geq \theta)}
\end{multline}
\end{small}



\subsection{Experiment Setting}
We run AlignScore, MiniCheck models along with \ours~ trained on the same backbone models for comparison. We select the same threshold used for benchmark \textbf{LLM-A{\small GGRE}F{\small ACT}} evaluation on the WiCE dataset for each model.

\subsection{Results and Analysis}

\begin{table}[t]  
\centering  
\resizebox{1.0\columnwidth}{!}{  
\small  
\begin{tabular}{lcc}  
\toprule  
\textbf{Model} & \textbf{Accuracy$_{\text{CoRe}}$} & \textbf{Precision$_{\text{CoRe}}$} \\  
\midrule  
AlignScore & 13.85 & 18.22 \\  
Minicheck-RBT & \textbf{24.42} & 34.32 \\  
FactCG-RBT & \textbf{24.42} & \textbf{54.04} \\  
\midrule  
Minicheck-DBT & 26.54 & 37.81 \\  
FactCG-DBT & \textbf{33.85} & \textbf{44.78} \\  
\midrule  
Minicheck-FT5 & 28.85 & 37.04 \\  
FactCG-FT5 & \textbf{32.31} & \textbf{45.53} \\  
\bottomrule  
\end{tabular}  
}  
\caption{\textbf{CoRe Evaluation Results.} Definition of two metircs can be found in section \ref{sec:core}.} 
\label{tb:dire}  
\end{table}

We show the CoRe evaluation results in Table \ref{tb:dire}. We observe consistent performance across the RBT, DBT and FT5 backbones. For RBT, FactCG achieves superior $\text{Precision}_{\text{CoRe}}$ while maintaining comparable $\text{Accuracy}_{\text{CoRe}}$ relative to MiniCheck. In the cases of FT5 and DBT, FactCG outperforms MiniCheck in terms of both $\text{Accuracy}_{\text{CoRe}}$ and $\text{Precision}_{\text{CoRe}}$. It shows that FactCG leverages more on the effective multi-hop reasoning chain across multi-hops than dataset artifacts. However, we still observe that half of the reasoning can be categorized as disconnected reasoning on both models. It demonstrates room for improvement in future trustworthy factuality models' reasoning process.


\section{Conclusion}
In this paper, we investigated the difference between state-of-the-art synthetic generated claims and real LLM-generated claims. To fill the gap, we proposed a new synthetic data generation approach, CG2C, that leverages the context graph to generate complex multi-hop claims without relying on LLMs to decide data labels. Our Fact Checker FactCG leveraging this generated data achieves state-of-the-art performance compared with models of similar parameter size and even outperforms GPT-4-o, which we used to construct the CG2C dataset.

\section{Limitations}
\subsection{Graph Extraction Quality}
We find it hard to define the granularity of the extracted triples. The simplest solution is to extract the $\langle noun, verb, noun \rangle$ triples from sentences, but information after prepositions and conjunctions may be lost. Long sentences and clauses make the task even more complicated. On the other hand, if we extract too many details, the graph will contain a lot of unnecessary edges and become too large to process efficiently. It is not easy to obtain optimal graphs for our use cases. Future research is needed to improve the graph extraction process.
In addition, although the graph extraction method can extract the relations between entities and help downstream models make inference, the extraction process itself cannot handle inference within the source document perfectly. For example, when we extract sub-graph for claims from the source documents' graphs in Section \ref{sec:poc}, we may observe the following two relations from the source document: "Shop A is open on Monday" and "Shop A is open on Tuesday". However, they cannot be matched exactly with "Shop A is open two days a week" in the claim. In this scenario, we have to rely on LLMs' inference ability to resolve the co-reference. 

\subsection{Data Generation Quantity}
Although our CG2C could support more scalable generation, we mainly extract few samples from one sub-graph to avoid duplication. Exploiting the graph to generate more samples and controlling the ratio of generated samples' hops could be a non-trivial topic to explore. 

\subsection{Chunking Methodology}
We follow previous work \cite{zha-etal-2023-alignscore, tang-etal-2024-minicheck} for the chunking methodology and fix the model input to a similar input token length. However, each scenario may have its own optimal chunk size. An area worth exploration for improving generalizable hallucination detection classifiers is developing intelligent methods to segment long documents into chunks based on topic flow.

\bibliography{main}

\appendix

\section{Appendix}
\label{sec:appendix}

\subsection{Benchmark Statistics} \label{sec: data_stats}
The statistics of \textbf{LLM-A{\small GGRE}F{\small ACT}} benchmark test set can be found in table~\ref{tab:dataset_sizes}. More details can be found in \citeauthor{tang-etal-2024-minicheck} (\citeyear{tang-etal-2024-minicheck}).

\begin{table}[h!]
\centering
\begin{tabular}{ll}
\toprule
\textbf{Dataset}         & \textbf{Size} \\
\midrule
AggreFact-CNN            & 558           \\
AggreFact-XSum           & 558           \\
TofuEval-MediaS          & 726           \\
TofuEval-MeetB           & 772           \\
Wice                     & 358           \\
Reveal                   & 1710          \\
ClaimVerify              & 1088          \\
FactCheck-GPT            & 1566          \\
ExpertQA                 & 3702          \\
Lfqa                     & 1911          \\
RAGTruth                 & 16371         \\
\bottomrule
\end{tabular}
\caption{\textbf{LLM-A{\small GGRE}F{\small ACT} benchmark test set size.}}
\label{tab:dataset_sizes}
\end{table}

\subsection{T-Test Analysis}
\begin{table*}[h]
\centering
\resizebox{\textwidth}{!}{
\begin{tabular}{lcccccccccccc}
\toprule
\multirow{2}{*}{\textbf{Measures}} & \multicolumn{2}{c}{\textbf{A{\small GGRE}F{\small ACT}}} & \multicolumn{2}{c}{\textbf{TofuEval}} & \multirow{2}{*}{\textbf{WiCE}} & \multirow{2}{*}{\textbf{REVEAL}} & \textbf{Claim} & \textbf{Fact} & \textbf{Expert} & \multirow{2}{*}{\textbf{LFQA}} & \textbf{RAG} & \multirow{2}{*}{\textbf{AVG}} \\ 
& \textbf{CNN} & \textbf{XSum} & \textbf{MediaS} & \textbf{MeetB} &  &  & \textbf{Verify} & \textbf{Check} & \textbf{QA} &  & \textbf{Truth} &  \\ 
\midrule
\textbf{Relative Improvement} & 10.1 & 0.0 & 4.8 & 3.3 & 4.4 & 2.6 & 2.7 & -0.8 & 0.6 & 2.4 & 3.7 & 3.1 \\ 
\textbf{p-value} & 3.9e-23 & 1.0e-01 & 2.0e-13 & 7.1e-08 & 3.1e-15 & 1.4e-12 & 9.0e-10 & 9.9e-01 & 4.5e-02 & 1.2e-08 & 2.3e-07 & 4.43e-40 \\ 
\textbf{p-value < 0.05} & True & False & True & True & True & True & True & False & True & True & True & True \\ 
\bottomrule
\end{tabular}
}
\caption{\textbf{T-test Comparison for FactCG-DBT and MiniCheck-DBT.} An apples-to-apples comparison with the same 
 DBT backbone. It allows us to validate the significance of the improvement presented in Table~\ref{tab:benchmark} and assess the effectiveness of our synthetic training data. Relative Improvement is defined as the BAcc (\%) delta between FactCG-DBT and MiniCheck-DBT.}
\label{tab:ttest1}
\end{table*}

\begin{table*}[h]
\centering
\resizebox{\textwidth}{!}{
\begin{tabular}{lcccccccccccc}
\toprule
\multirow{2}{*}{\textbf{Measures}} & \multicolumn{2}{c}{\textbf{A{\small GGRE}F{\small ACT}}} & \multicolumn{2}{c}{\textbf{TofuEval}} & \multirow{2}{*}{\textbf{WiCE}} & \multirow{2}{*}{\textbf{REVEAL}} & \textbf{Claim} & \textbf{Fact} & \textbf{Expert} & \multirow{2}{*}{\textbf{LFQA}} & \textbf{RAG} & \multirow{2}{*}{\textbf{AVG}} \\ 
 & \textbf{CNN} & \textbf{XSum} & \textbf{MediaS} & \textbf{MeetB} & & & \textbf{Verify} & \textbf{Check} & \textbf{QA} & & \textbf{Truth} & \\ 
\midrule
\textbf{Relative Improvement} & 3.7 & -0.1 & 0.6 & -2.3 & 4.9 & 3.4 & 3.6 & -1.7 & 0.8 & 0.9 & 4.1 & 1.7 \\ 
\textbf{p-value} & 0.6e-03 & 2.6e-01 & 3.6e-03 & 9.9e-01 & 9.7e-18 & 4.6e-15 & 5.9e-11 & 1.0e-05 & 4.4e-03 & 3.2e-03 & 9.5e-08 & 8.0e-14 \\ 
\textbf{p-value < 0.05} & True & False & True & False & True & True & True & False & True & True & True & True \\ 
\bottomrule
\end{tabular}
}
\caption{\textbf{T-test Comparison for FactCG-DBT and MiniCheck-FT5.} We validate the significance of the improvement for FactCG-DBT over the best performing fact-checker on LLM-Aggrefact. Relative Improvement is defined as the BAcc (\%) delta, representing the difference between FactCG-DBT and MiniCheck-FT5.}
\label{tab:ttest2}
\end{table*}
We conduct a paired bootstrap test with 100 runs and a p-value < 0.05. The bootstrap sample size is set to 150. Our experiment aims to reject the null hypothesis that the observed differences between methods are due to random chance.

We first compare our FactCG-DBT with MiniCheck-DBT on synthetic data generation as they share the same model backbone. 
We set $H_0$ as the performance on BAcc of FactCG-DBT is no better than that of MiniCheck-DBT. We show results in Table~\ref{tab:ttest1}.

On average, FactCG-DBT demonstrates a significant improvement in performance (BAcc) compared to MiniCheck-DBT with p-value less than 0.05. Therefore we reject the null hypothesis that FactCG-DBT performs no better than MiniCheck-DBT. Notably, in datasets where FactCG-DBT exhibits a relative improvement, the p-values are consistently below 0.05 and align with the observed BAcc gains. This demonstrates the effectiveness of our synthetic training data.

We additionally compare FactCG-DBT with MiniCheck-FT5, which is the best-performing fact-checker (<0.8B) on LLM-AggreFact, to prove the significance of our improvement. We set $H_0$ as the performance on BAcc of FactCG-DBT is no better than that of MiniCheck-FT5. We show results in Table~\ref{tab:ttest2}.

Similar to the above experiment, the averaged performance shows a statistically significant improvement, providing strong evidence to reject the null hypothesis that FactCG-DBT performs no better than MiniCheck-FT5. In datasets where FactCG-DBT demonstrates improvement, the p-values are consistently below 0.05, confirming the statistical significance of these gains.

In conclusion, the results of both t-tests support our assertion of the statistically significant improvement of FactCG-DBT and align with the BAcc gains in Table~\ref{tab:benchmark}.

\subsection{Results without Threshold Tuning} \label{sec: threshold0.5}
We use threshold 0.5 for our \ours~ model. We shared our results without specific task threshold tuning on Table \ref{tab:benchmark-fixedthreshold}. Our \ours-DBT achieves the highest balanced accuracy compared to MiniCheck.

\begin{table*}[ht]
\centering
\caption{\textbf{LLM-A{\small GGRE}F{\small ACT} Results without threshold tuning.} }
\resizebox{2\columnwidth}{!}{
\begin{tabular}{lcccccccccccc}
\hline
\multirow{2}{*}{\textbf{Model}} & \multicolumn{2}{c}{\textbf{A{\small GGRE}F{\small ACT}}} & \multicolumn{2}{c}{\textbf{TofuEval}} & \multirow{2}{*}{\textbf{WiCE}} & \multirow{2}{*}{\textbf{REVEAL}} & \textbf{Claim} & \textbf{Fact} & \textbf{Expert} & \multirow{2}{*}{\textbf{LFQA}} & \textbf{RAG} & \multirow{2}{*}{\textbf{AVG}} \\
 \textbf{} & \textbf{CNN} & \textbf{XSum} & \textbf{MediaS} & \textbf{MeetB} & \textbf{} & \textbf{} & \textbf{Verify} & \textbf{Check} & \textbf{QA} & \textbf{} & \textbf{Truth} & \textbf{} \\
\midrule
GPT-4o-2024-05-13 & 68.1 &  76.8 & 71.4 & 79.8 & 78.5 & 86.5 & 69.0 & 77.5 & 59.6 & 83.6 &  84.3 & 75.9 \\
\midrule
\midrule
Summac-CV & 65.2 & 54.5 & 63.7 & 62.8 & 54.3 & 67.7 & 70.9 & 53.4 & 54.9 & 62.1 & 61.7 & 61.0 \\
Summac-ZS & 51.1 & 61.5 & 69.5 & 71.0 & 62.8 & 85.3 & 69.7 & \cellcolor{darkgreen}75.2 & 55.2 & 77.6 & 65.6 & 67.7  \\ 
AlignScore & 52.4 & 71.4 & 69.2 & 72.6 & 66.0 & 85.3 & 69.6 & 74.3 & 58.3 & 84.5 & 71.7 & 70.5 \\
MiniCheck-RBT & 63.7 & 70.8 & 71.9 & \cellcolor{lightgreen}75.9 & 67.6 & \cellcolor{lightgreen}88.8 & 77.4 & 73.3 & 57.4 & 84.4 & 77.2 & 73.5 \\
MiniCheck-DBT & 64.2 & 71.0 & 69.3 & 72.7 & 69.4 & 87.3 & 75.6 & 73.0 & 58.9 & 83.9 & 78.8 & 73.1 \\
MiniCheck-FT5 & 69.9 & 74.3\cellcolor{darkgreen} & \cellcolor{lightgreen}73.6 & \cellcolor{darkgreen}77.3 & 72.2 & 86.2 & 74.6 & \cellcolor{lightgreen}74.7 & \cellcolor{lightgreen}59.0 & 85.2 & 78.0 & \cellcolor{lightgreen}75.0 \\
\midrule
\ours-RBT & \cellcolor{lightgreen}70.1 & 67.0 & 70.0 & 73.4 & 64.9 & \cellcolor{darkgreen}89.7 & 77.8 & 72.4 & 57.7 & 86.5 & 78.2 & 73.4 \\
\ours-DBT & \cellcolor{darkgreen}70.2 & \cellcolor{lightgreen}73.9 & 72.3 & 74.3 & \cellcolor{darkgreen}74.2 & 88.4 & \cellcolor{lightgreen}78.5 & 72.1 & \cellcolor{darkgreen}59.1 & \cellcolor{lightgreen}86.7 & \cellcolor{darkgreen}82.3 & \cellcolor{darkgreen}75.6
 \\
\ours-FT5 & 62.8 & 73.2 & \cellcolor{darkgreen}74.3 & 75.5 & \cellcolor{lightgreen}72.8 & \cellcolor{darkgreen}89.7 & \cellcolor{darkgreen}78.6 & 73.0 & 58.3 & \cellcolor{darkgreen}88.2 & \cellcolor{lightgreen}78.9 & \cellcolor{lightgreen}75.0 \\
\bottomrule
\end{tabular}
}
\label{tab:benchmark-fixedthreshold}
\end{table*}

\subsection{Model Size and Computational Cost}

We conduct our experiment on Nvidia Quadro RTX 8000 and train all FactCG model backbones with 4 GPUs and inference with 1 GPU. Training costs and the inference costs (including threshold selection) are shown below:

For our 0.4B size \ours-RBT, the training takes approximately 7 minutes and the inference takes about about 65 minutes to complete.

For our 0.4B size \ours-DBT, the training costs approximately 18 minutes and the inference takes about 103 minutes to complete.

For our 0.8B size \ours-T5, the training takes approximately 32 minutes and the inference takes about 238 minutes to complete.

\subsection{Scientific Artifacts}
We list the licenses used in this paper: pytorch-lightning (Apache License 2.0), PyTorch (BSD-3),  Huggingface Transformers (Apache License 2.0), MiniCheck (Apache License Version 2.0), AlignScore (MIT License), RoBERTa-Large-MNLI (MIT License), hotpotqa (Apache License 2.0), musique (Creative Commons Attribution 4.0 International),  OpenAI (Term of use\footnote{https://openai.com/policies/row-terms-of-use/}). We adhere to the intended use of all the existing artifacts mentioned in this paper.

\subsection{Packages}
We employed the following package to conduct our experiment: pytorch-lightning v2.4.0, PyTorch v2.3.0+cu121, and transformers v4.44.2.

\begin{table*}[ht]
\centering
\setlength{\arrayrulewidth}{1pt}
\caption{\ours-DBT and \ours-RBT Instruction Template} \label{tab: instruction template}
\begin{tabular}{p{2\columnwidth}}
\hline
\\

<DOCUMENT>.\\
\\
Choose your answer: based on the paragraph above can we conclude that "<CLAIM>"? \\
\\
OPTIONS: \\
- Yes \\
- No \\
I think the answer is \\
\\
\hline
\end{tabular}
\end{table*}

\subsection{\ours-DBT and \ours-RBT Instruction Template}
We show \ours-DBT and \ours-RBT Instruction Template in Table \ref{tab: instruction template}

\subsection{Prompts and GPT Usage}
We use GPT-4-o to create CG2C datasets. 
For each positive and negative sample pair generated from HotpotQA, we run four GPT calls in order to generate a claim (Table \ref{tab: prompt: qa2c}), a context graph (Table \ref{tab: prompt: CG from QA}), a context sub-graph (Table \ref{tab: prompt: sub CG}), and a negative sample's document (Table \ref{tab: prompt: relatioin removal}).
For each positive sample in CG2C-MHQA generated from Musique, we run one GPT call to generate a claim (Table \ref{tab: prompt: qa2c}).
To construct CG2C-Doc dataset, we run one GPT call for each unique document (around 1300) in MiniCheck's D2C dataset to extract the context graph (Table \ref{tab: prompt: CG from Doc}). In addition, for each positive and negative sample pair we generated, we call GPT two more times to generate the claim (Table \ref{tab: prompt: claim generation from graph}) and negative sample's document (Table \ref{tab: prompt: relatioin removal}).


\begin{table*}[ht!]
\centering
\caption{Prompt: Context Graph Extraction for CG2C-MHQA}\label{tab: prompt: CG from QA}
\setlength{\arrayrulewidth}{1pt}
\begin{tabular}{p{2\columnwidth}}
\hline

You are a helpful assistant.\\
\\
Extract content graph with sentences in forms of triples (entity, relation, entity) based only on the provided sentences. Provide the triples only.\\
\\
Examples:\\
\\
Provided Sentences:
\begin{itemize}
\setlength\itemsep{-0.2em}
\item Scott Derrickson (born July 16, 1966) is an American director, screenwriter and producer.
\item Edward Davis Wood Jr. (October 10, 1924 – December 10, 1978) was an American filmmaker, actor, writer, producer, and director.
\end{itemize}
\\
Triples in Provided Sentences:
\begin{itemize}
\setlength\itemsep{-0.2em}
\item (Scott Derrickson, born on, July 16, 1966)
\item (Scott Derrickson, is, American)
\item (Scott Derrickson, is, director)
\item (Scott Derrickson, is, screenwriter)
\item (Scott Derrickson, is, producer)
\item (Edward Davis Wood Jr., born on, October 10, 1924)
\item (Edward Davis Wood Jr., died on, December 10, 1978)
\item (Edward Davis Wood Jr., was, American)
\item (Edward Davis Wood Jr., was, filmmaker)
\item (Edward Davis Wood Jr., was, actor)
\item (Edward Davis Wood Jr., was, writer)
\item (Edward Davis Wood Jr., was, producer)
\item (Edward Davis Wood Jr., was, director)
\end{itemize}
Your turn:\\
\\
Provided Sentences:\\
<SENTENCES>\\
\\
Triples in Provided Sentences:\\

\hline
\end{tabular}
\end{table*}

\begin{table*}[ht!]
\centering
\caption{Prompt: Context Graph Extraction for CG2C-Doc}\label{tab: prompt: CG from Doc}
\setlength{\arrayrulewidth}{1pt}
\begin{tabular}{p{2\columnwidth}}
\hline
You are a helpful assistant.\\
\\
Given a news article, go over every sentence and extract triples in forms of (entity \{tuple\_delimiter\} entity \{tuple\_delimiter\} a short description about the relation between two entities). Group the triples with the same entity. Separate groups of triples using \{group\_delimiter\}.\\
\\
Examples:\\
\\
Provided Sentences: \\
Hunt also said the government needs to reform the welfare system to get more people back to work. The number of people not in the workforce for physical or mental health reasons has soared since the pandemic. Ken Clarke, a former Conservative Treasury chief, said cutting inheritance tax "might appeal to the Conservative right, but it leaves them open to the most appalling criticisms when inflation and the state of affairs is making poorer people in this country very vulnerable indeed." "I'm not sure that the economic and financial state of the country justifies it." \\
\\
Groups of Triples in Provided Sentences:
\begin{itemize}
\setlength\itemsep{-0.2em}
\item Hunt \{tuple\_delimiter\} government \{tuple\_delimiter\} Hunt said something about the government.
\item government \{tuple\_delimiter\} welfare system \{tuple\_delimiter\} government need to reform welfare system.
\item welfare system \{tuple\_delimiter\} people \{tuple\_delimiter\} reformed welfare system can get people back to work.
\item people \{tuple\_delimiter\} physical or mental health reasons \{tuple\_delimiter\} people has soared due to physical or mental health reasons
\item \{group\_delimiter\}
\item Ken Clarke \{tuple\_delimiter\} former Conservative Treasury chief \{tuple\_delimiter\} Ken Clarke is former Conservative Treasury chief
\item Ken Clarke \{tuple\_delimiter\} cutting inheritance tax \{tuple\_delimiter\} Ken Clarke said something about cutting inheritance tax
\item cutting inheritance tax \{tuple\_delimiter\} Conservative right \{tuple\_delimiter\} cutting inheritance tax appeal to the Conservative right
\item cutting inheritance tax \{tuple\_delimiter\} criticisms \{tuple\_delimiter\} cutting inheritance tax leaves open to criticisms
\item cutting inheritance tax \{tuple\_delimiter\} inflation and the state of affairs \{tuple\_delimiter\} cutting inheritance tax cause inflation and the state of affairs
\item inflation and the state of affairs \{tuple\_delimiter\} poorer people \{tuple\_delimiter\} inflation and the state of affairs make poorer people vulnerable
\item economic and financial state of the country \{tuple\_delimiter\} cutting inheritance tax \{tuple\_delimiter\} economic and financial state of the country might not justify cutting inheritance tax.
\end{itemize}

Your turn:\\
\\
Provided Sentences:\\
<SENTENCES>\\
\\
Groups of Triples in Provided Sentences:\\
\hline
\end{tabular}
\end{table*}

\begin{table*}[ht!]
\centering
\caption{Prompt: Multi-hop Reasoning Chain Mapping for Claim Sub-graph Extraction}\label{tab: prompt: sub CG}
\setlength{\arrayrulewidth}{1pt}
\begin{tabular}{p{2\columnwidth}}
\hline

You are a helpful assistant.\\
\\
Find the triplets in forms of (entity \{tuple\_delimiter\} entity \{tuple\_delimiter\} a short description about the relation between two entities) that exist in the provided sentence with the provided triplets only. Directly output the Existed Triplets in Provided Sentences.\\
\\
Examples:\\
Provided Triples:
\begin{itemize}
\vspace{-0.5em}
\setlength\itemsep{-0.2em}
\item (Scott Derrickson \{tuple\_delimiter\} July 16, 1966 \{tuple\_delimiter\} Scott Derrickson was born on July 16, 1966)
\item (Scott Derrickson \{tuple\_delimiter\} American \{tuple\_delimiter\} Scott Derrickson is an American)
\item (Scott Derrickson \{tuple\_delimiter\} director \{tuple\_delimiter\} Scott Derrickson is a director)
\item (Scott Derrickson \{tuple\_delimiter\} screenwriter \{tuple\_delimiter\} Scott Derrickson is a screenwriter)
\item (Scott Derrickson \{tuple\_delimiter\} producer \{tuple\_delimiter\} Scott Derrickson is a producer)
\item (Edward Davis Wood Jr. \{tuple\_delimiter\} October 10, 1924 \{tuple\_delimiter\} Edward Davis Wood Jr. was born on October 10, 1924)
\item (Edward Davis Wood Jr. \{tuple\_delimiter\} December 10, 1978 \{tuple\_delimiter\} Edward Davis Wood Jr. was died on December 10, 1978)
\item (Edward Davis Wood Jr. \{tuple\_delimiter\} American \{tuple\_delimiter\} Edward Davis Wood Jr. was an American)
\item (Edward Davis Wood Jr. \{tuple\_delimiter\} filmmaker \{tuple\_delimiter\} Edward Davis Wood Jr. was a filmmaker)
\item (Edward Davis Wood Jr. \{tuple\_delimiter\} actor \{tuple\_delimiter\} Edward Davis Wood Jr. was an actor)
\item (Edward Davis Wood Jr. \{tuple\_delimiter\} writer \{tuple\_delimiter\} Edward Davis Wood Jr. was a writer)
\end{itemize}
\\
Provided Sentences:\\
Scott Derrickson and Ed Wood were both of the same nationality.\\
\\
Existed Triplets in Provided Sentences:
\begin{itemize}
\vspace{-0.5em}
\setlength\itemsep{-0.2em}
\item (Scott Derrickson \{tuple\_delimiter\} American \{tuple\_delimiter\} Scott Derrickson is an American)
\item (Edward Davis Wood Jr. \{tuple\_delimiter\} American \{tuple\_delimiter\} Edward Davis Wood Jr. was an American)
\end{itemize}
\\
Your turn:\\
\\
Provided Triples:\\
<TRIPLES>\\
\\
Provided Sentences:\\
<SENTENCES>\\
\\
Existed Triplets in Provided Sentences:\\

\hline
\end{tabular}
\end{table*}

\begin{table*}[ht!]
\centering
\caption{QA to Declarative Statement Prompt}\label{tab: prompt: qa2c}
\setlength{\arrayrulewidth}{1pt}
\begin{tabular}{p{2\columnwidth}}
\hline

You are a helpful assistant.\\
\\
Question: <QUESTION>\\
\\
Answer: <ANSWER>\\
\\
Convert the question answer pair into a single declarative sentence.\\
\\
Single declarative sentence:\\

\hline
\end{tabular}
\end{table*}

\begin{table*}[ht!]
\centering
\caption{Prompt: Document Relation Removal}\label{tab: prompt: relatioin removal}
\setlength{\arrayrulewidth}{1pt}
\begin{tabular}{p{2\columnwidth}}
\hline

You are a helpful assistant.\\
\\
Remove the relation between two provided entities in below provided sentences with minimal changes. Provide the rewrite sentences only.\\
\\
Examples:\\
\\
Provided Entities:\\
India, Area\\
\\
Provided Sentences:\\
Indogrammodes is a genus of moths of the Crambidae family. It contains only one species, Indogrammodes pectinicornalis, which is found in India. India, officially the Republic of India ("Bhārat Gaṇarājya"), is a country in South Asia. It is the seventh-largest country by area, the second-most populous country (with over 1.2 billion people), and the most populous democracy in the world.\\
\\
Rewrite Sentences with relationship between provided two entites removed:\\
Indogrammodes is a genus of moths of the Crambidae family. It contains only one species, Indogrammodes pectinicornalis, which is found in India. India, officially the Republic of India ("Bhārat Gaṇarājya"), is a country in South Asia. It is the second-most populous country (with over 1.2 billion people) and the most populous democracy in the world.\\
\\
Your turn:\\
\\
Provided Entities:\\
<ENTITIES>\\
\\
Provided Sentences:\\
<SENTENCES>\\
\\
Rewrite Sentences with Relationship Between Provided Two Entites Removed:\\

\hline
\end{tabular}
\end{table*}

\clearpage
\begin{table*}[t]
\centering
\caption{Prompt: Claim Generation from Graph}\label{tab: prompt: claim generation from graph}
\setlength{\arrayrulewidth}{1pt}
\begin{tabular}{p{2\columnwidth}}
\hline

You are a helpful assistant.\\
\\
Given a news article, write a short sentence, which must include the following entities: <ENTITIES>\\
\\
Provided Article:\\
<SENTENCES>\\
\\
Your Claim:\\

\hline
\end{tabular}
\end{table*}

\end{document}

%% file: math_commands.tex
\usepackage{amsmath,amsfonts,bm}

\def\eqref#1{equation~\ref{#1}}

\def\1{\bm{1}}

\DeclareMathAlphabet{\mathsfit}{\encodingdefault}{\sfdefault}{m}{sl}
\SetMathAlphabet{\mathsfit}{bold}{\encodingdefault}{\sfdefault}{bx}{n}

\def\gM{{\mathcal{M}}}



%% file: macro.tex
\definecolor{lightgreen}{RGB}{201,242,155}
\definecolor{darkgreen}{RGB}{0,170,136}
\definecolor{na}{gray}{0.9}
\newcommand{\start}[1]{\vspace{.3mm}\noindent{{\bf #1}.}}